\documentclass{article}
\usepackage{fullpage}
\usepackage{graphicx}
\begin{document}
\begin{center}
\huge Understanding ACT-R -- an Outsider's Perspective
\normalsize

Jacob Whitehill, \begin{tt}jake@mplab.ucsd.edu\end{tt}
\end{center}

\abstract{
	The ACT-R theory of cognition developed by John Anderson and colleagues endeavors to explain how humans recall chunks of information and how they solve problems. ACT-R also serves as a theoretical basis for ``cognitive tutors'', i.e., automatic tutoring systems that help students learn mathematics, computer programming, and other subjects. The official ACT-R definition is distributed across a large body of literature spanning many articles and monographs, and hence it is difficult for an ``outsider'' to learn the most important aspects of the theory. This paper aims to provide a tutorial to the core components of the ACT-R theory.
}

\section{Introduction}
\label{sec:introduction}
The Adaptive Character of Thought - Rational (ACT-R) is a theory of cognition developed principally by
John Anderson at Carnegie-Mellon University \cite{Anderson1993}. ACT-R models how humans recall ``chunks'' of
information from memory and how they solve problems by breaking them down into subgoals and applying
knowledge from working memory as needed.

ACT-R is the latest in a series of cognitive models in the ACT family; it succeeds ACT, ACTE, and ACT*. With ACT*
\cite{Anderson1983}, Anderson endeavored to describe the human memory and reasoning faculties \emph{mechanistically} -- i.e.,
to describe the mechanisms through which memory recall and cognition take place --
by describing how goals are broken down into subgoals using ``production rules.'' Later, in 1990, Anderson took a step back 
from the existing ACT* and asked how cognitive processes such as memory, categorization, causal inference, and
problem solving could be cast as \emph{optimal solutions} to the tasks that humans commonly encounter.
The underlying belief was that humans, from an evolutionary perspective, represent a local maximum of adaptedness to their environment; therefore,
their basic memory retrieval and decision-making processes should also be somehow optimal. Optimality 
severely constrains the possible mechanisms of human cognition and thus helps to reduce the search space for the true underlying mechanism.

With his new-found optimization approach, Anderson returned in 1993 to the ACT production rule-based framework and revised ACT* to create ACT-R. The ``R''
(rational) implies that the human mind is behaving ``rationally,'' in the sense that it wishes to solve problems to maximize reward and
minimize cost. This formulation of cognition (with costs and rewards) is somewhat reminiscent of stochastic optimal control theory.

Unfortunately, in contrast to the precisely defined mathematics of stochastic optimal control, the ACT-R literature is marred by a lack of
specificity and consistency. The ACT-R definition is distributed across a vast collection of journal articles and monographs,
and no single text is sufficient to provide a complete definition.
Important parts of the ACT-R definition vary from source to source, with no explanation as to why the
change was made, or even an acknowledgement that the change had occurred at all. (An example of this is the decay rate of a learning
event for estimating its effect on activation.) Mathematically precise terminology such as ``log odds of X''
(for some event X) is used injudiciously, without any proof that the associated quantity equals what it should. The reader must
sometimes guess what the author meant to say in order for the equations to hold true.

This tutorial endeavors to describe clearly the main ideas of ACT-R from the top-level decision-making processes down to the
level of strengthening the presence of knowledge chunks in memory.  The goal of the document is to provide clarity
where the original ACT-R literature was vague or inconsistent, and to summarize in one relatively short document the ``complete picture''
of ACT-R (or at least the main points) that are scattered throughout the vast ACT-R literature corpus. 

\subsection{Roadmap}
We first introduce the crucial distinction between \emph{declarative knowledge } and \emph{procedural knowledge} in Section \ref{sec:decl_proc}.
The document then proceeds in a top-down fashion: Under the assumption that the agent (a human, or possibly a computer)
already has all of the knowledge he/she needs, we examine in Section \ref{sec:decision_making} how the decision-making process is made
on a rational basis under ACT-R. In particular, we describe
the mechanism by which a particular ``production rule,'' corresponding to the ``actions'' of ACT-R, is chosen out of many possible alternatives.
In Section \ref{sec:learning}, we remove the assumption that knowledge is already available, and describe the ACT-R processes by which new knowledge
is acquired. This includes both the creation of new memories, as well as the strengthening (and decay)
of existing ones. Finally, in Sections \ref{sec:spacing} and \ref{sec:power_laws}, we discuss how ACT-R partially models
the Spacing Effect and the Power Laws of Learning/Forgetting.

\section{Declarative versus Procedural Knowledge}
\label{sec:decl_proc}
Under ACT-R, human knowledge is divided into two disjoint but related sets of knowledge -- \emph{declarative} and \emph{procedural}.
\emph{Declarative} knowledge comprises many \emph{knowledge chunks}, which are the current set of facts that are known and goals that are active.
Two examples of chunks are ``The bank is closed on Sundays,'' and ``The current goal is to run up a hill.'' Notice that each chunk may 
refer to other chunks. For instance, our first example chunk refers to the concepts of ``bank,'' ``closed,'' and ``Sunday,'' which presumably
are themselves all chunks in their own right. When a chunk $i$ refers to, or is referred to
by, another chunk $j$, then chunk $i$ is said to be \emph{connected} to chunk $j$. This relationship is not clearly defined in ACT-R -- for instance,
whether the relationship is always symmetrical, or whether it can be reflexive (i.e., a chunk referring to itself in recursive fashion), is
not specified.

\emph{Procedural} knowledge is the
set of \emph{production rules} -- if/then statements that specify how a particular goal can be achieved when a specified pre-condition is met -- 
that the agent currently knows. A production rule might state, for instance, ``If I am hungry, then eat.'' For the domain of intelligent
tutoring systems, for which ACT* and ACT-R were partly conceived, a more typical rule might be, ``If the goal is to prove triangle similarity, then prove
that any two pairs of angles are congruent.''

The human memory contains many declarative knowledge chunks and production rules. At any point in time, when a person is trying to
complete some task, a production rule, indicating the next step to take in order to solve the problem, may ``fire'' if the rule's pre-condition,
which is a conjunction of logical propositions that must hold true according to the current state of declarative memory, is fulfilled.
Since the currently available set of knowledge chunks may fulfill
the pre-conditions of multiple production rules, a competition exists among production rules to select the one that will actually fire. (This
competition will be described later.) Whichever rule ends up firing may result either in the goal being achieved, or in the creation of new knowledge
chunks in working memory, which may then trigger more production rules, and so on.

\subsection{Example}
Let us consider a concrete setting in order to make the above ideas more concrete. ACT-R most readily lends itself to domains in which tasks can be
decomposed into well-defined component operations. Suppose the current task the person is working on is to add two multi-digit numbers.
Table \ref{tbl:example_productions} proposes a plausible set of productions with which this problem can be solved.
\begin{table}
\begin{center}
\begin{tabular}{c|l|l}
\multicolumn{3}{c}{\bf Production Rules}\\\hline
P1 & If   & the goal is to add two numbers\\
   & Then & push a subgoal to process each of the columns with $k=0$.\\\hline
P2 & If   & the goal is to process each of the columns\\
   & and  & the last processed column was $k$ (counted from the right)\\
   & and  & column $k+1$ exists\\
   & Then & push a subgoal to process column $k+1$.\\\hline
P3 & If   & the goal is to process each of the columns\\
   & and  & the last processed column was $k$ (counted from the right)\\
   & and  & column $k+1$ does not exist\\
   & Then & pop subgoal.\\\hline
P4 & If   & the goal is to process column $k$\\
   & and  & the digits in the column are $x_k$, $y_k$, and carry $c_k$,\\
   & and  & $z_k=x_k+y_k+c_k < 10$,\\
   & Then & write $z_k$ below column $k$; pop subgoal.\\\hline
P5 & If   & the goal is to process column $k$\\
   & and  & the digits in the column are $x_k$, $y_k$, and carry $c_k$,\\
   & and  & $z_k=x_k+y_k+c_k \geq 10$,\\
   & Then & write the one's digit of $z_k$ below column $k$; set $c_{k+1}$ to 1\\
   &      & (adding a column of $0$'s as necessary); pop subgoal.\\\hline
P6 & If   & the goal is to add two numbers\\
   & and  & all the columns have been processed\\
   & Then & pop the goal.
\end{tabular}
\end{center}
\caption{Set of production rules sufficient to perform base-10 addition.}
\label{tbl:example_productions}
\end{table}
Let us assume for now that the agent (a math student, presumably) already possesses the productions given in Table \ref{tbl:example_productions})
in his/her procedural memory, and also that he/she can perform single-digit addition. This latter assumption could be fulfilled in two ways: either
the student has memorized these basic addition facts by rote, in which case they would be stored in declarative memory; or he/she
knows some counting procedure (e.g., using the fingers) to implement simple addition. This procedure would correspond to another production
rule in its own right, which for brevity we have omitted.

Given the current goal to add two multi-digit numbers -- in the present example, $36$ and $23$ -- the student's mind, under the ACT-R
model, implicitly matches the available declarative memory chunks to the pre-condition of all productions. In the example shown,
based on the contents of declarative memory, there is always only one possible production rule that can be applied at each
step, but ACT-R allows for more general scenarios and specifies how to select from multiple production rules
whose pre-conditions are all fulfilled.

We briefly describe the execution of the first few steps. At the onset, the only contents of memory relevant to the addition task is that
the goal is to add two numbers. This ``chunk'' of information -- a goal chunk, in this case -- matches the pre-condition
of only a single production, P1. After some small amount of memory retrieval time (latency of production matching will be
described in Section \ref{sec:learning}), the production will ``fire,'' meaning that its action -- embedded in the ``Then''
part of the production rule -- will be executed. In the case of P1, a new chunk is written to declarative memory storing the information
\begin{tt}k=0\end{tt}, and a subgoal ``Process columns'' is pushed onto the ``goal stack.''

The above paragraph describes one complete step of execution within the decision-making framework of ACT-R. At this point, a new goal chunk sits
at the top of the goal stack, and declarative memory contains a new chunk (\begin{tt}k=0\end{tt}). The process then proceeds as before -- the set
of declarative memory chunks is matched to all the productions' pre-conditions, causing one of these productions to fire. At later points of
execution, instead of ``pushing'' a subgoal onto the goal stack, a ``subroutine'' of execution will have completed and a production will
instead ``pop'' a goal chunk off the stack.

It should be noted that ACT-R only supports a \emph{single} goal stack; how multiple conflicting or completing goals (which are ubiquitous
in real life) are handled is not explored. A theory of \emph{goal pre-emption} -- the process whereby a more important goal interrupts
decision-making on behalf of a less important one -- is also not developed.

\begin{figure}
\begin{center}
\includegraphics[width=5.75in]{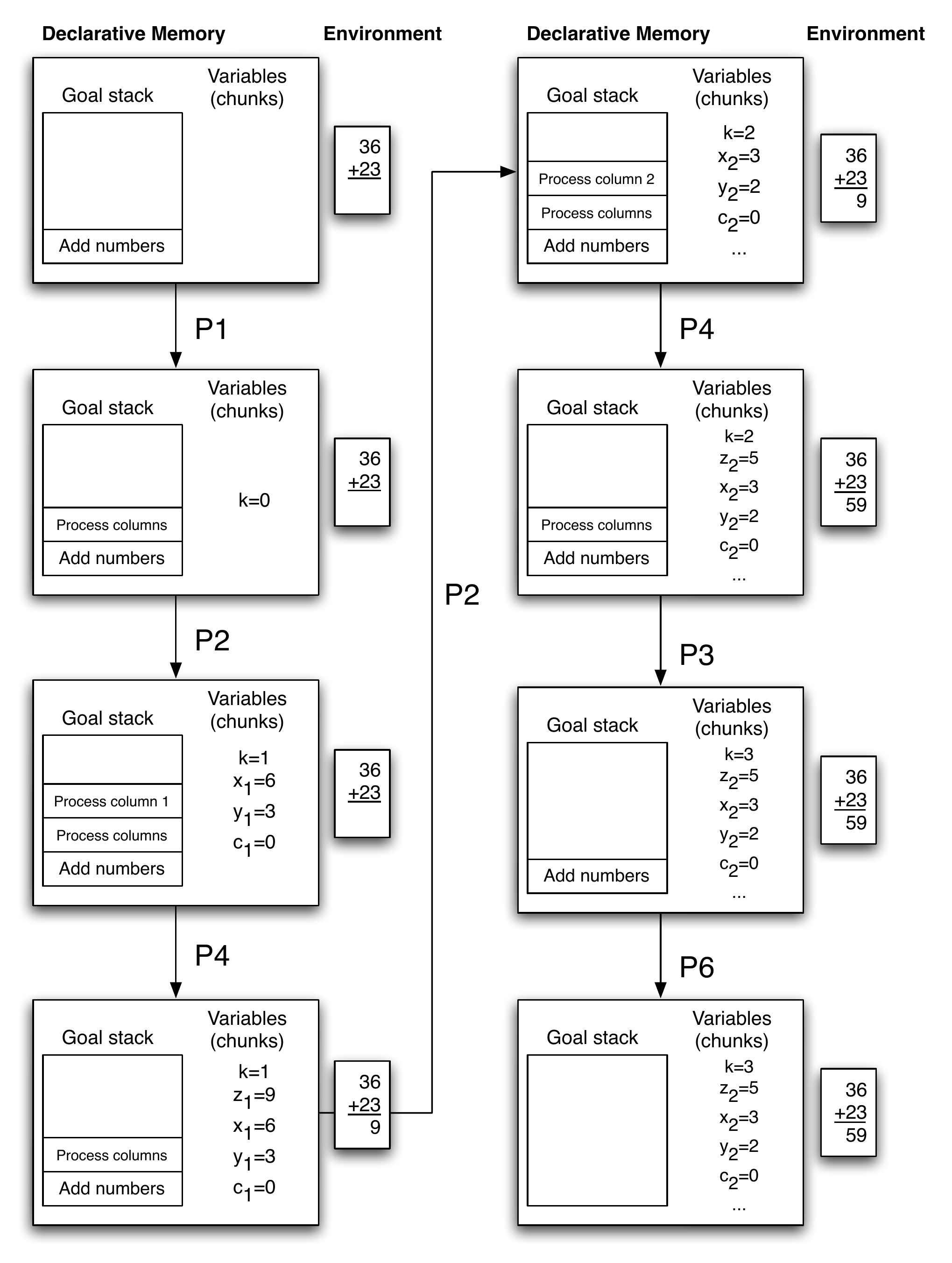}
\caption{An example trace through an addition problem using the productions in Table \ref{tbl:example_productions}.}
\label{fig:example_productions}
\end{center}
\end{figure}

The entire production queue from start to finish for the given addition problem ($36+23$) is shown in Figure \ref{fig:example_productions},
along with the contents of working memory and the state of the environment at each time-step. In the next section, we describe the
general decision-making process under ACT-R in which multiple productions must compete for execution.

\section{Decision Making in ACT-R}
\label{sec:decision_making}
At any moment in time, the knowledge chunks in declarative memory may fulfill, or \emph{match}, a part of the pre-condition of a production rule (or of
multiple production rules). 
When all parts of a production rule's pre-condition are fulfilled (recall that a pre-condition is a logical conjunction), then
production rule itself is said to \emph{match}/be \emph{matched}.
Given multiple production rules that are matched, a decision
must still be made as to which production rule will actually fire (be executed). This decision is made on the basis of the expected value
$V$ of each production that matches, as well as the time $t$ when the matching occurs. The expected value $V$ is learned by the agent through experience, and
the time of match $t$ is dictated by the \emph{latency} of matching, which decreases with learning. We will discuss learning in Section \ref{sec:learning};
for now, however, assume that the time of matching $t$, and the expected value $V$, of each production is already known by the agent.

\subsection{Flowchart}
\begin{figure}
\begin{center}
\includegraphics[width=5.75in]{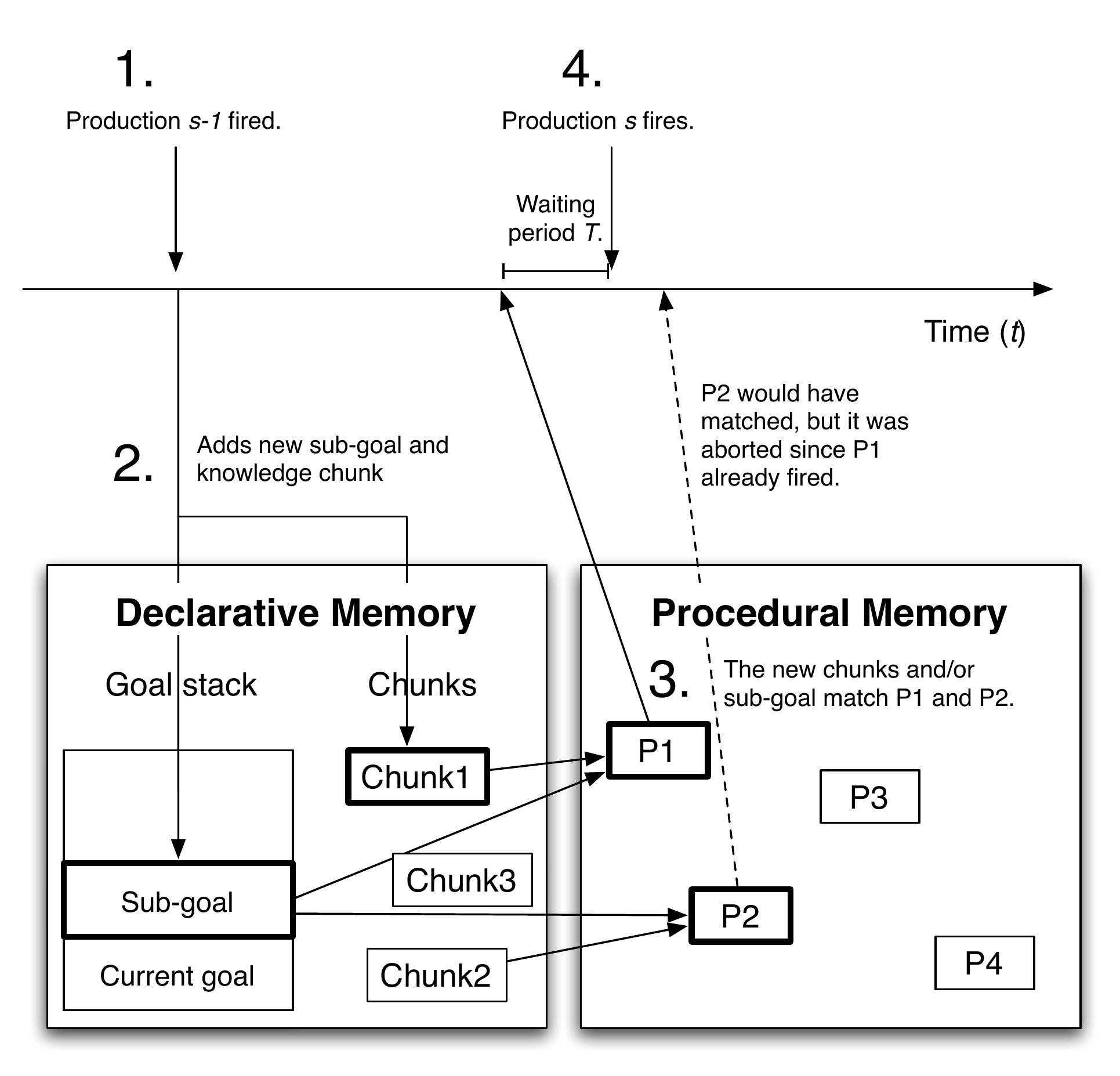}
\end{center}
\caption{Flowchart of how an ACT-R agent decides which production to select (fire).}
\label{fig:decision_process}
\end{figure}
The decision process of choosing which production to fire is shown in Figure \ref{fig:decision_process}.
Suppose production $s_i$ had matched at time $t$ and that the expected value of $s_i$ is $V_i=p_iG-C_i$, where $G$ is
the reward of achieving the goal, $p$ is the probability of achieving the goal if production $s_i$ fires, and $C_i$ is
the expected cost (both immediate and future) of $s_i$. Note that $C_i$ refers to ``real-world'' costs of performing the production's action (e.g.,
the cost of gasoline if, for instance, driving a car is involved),
not memory retrieval costs (which will be discussed momentarily). According to ACT-R, the agent must decide whether to choose (fire)
production $s_i$, or instead to wait for a better production $s_j$ (with higher value $V_j$). This decision is
made by assessing whether the expected gain of waiting exceeds the expected cost -- we call this $\tau$.

In ACT-R, the cost of waiting is modelled by a single constant, designed to reflect the memory retrieval cost of matching another production
$s_j$ in the future. A constant valued cost can only approximate the true cost for two reasons: first, several higher-valued productions
may in fact end up matching (if each is better than the last) and thus being retrieved, and hence the cost should actually vary 
with the number of productions whose value exceeds $V_i$. Second, each production rule may refer to a different number of declarative
memory chunks; presumably, retrieving each of these takes some energy as well. Nevertheless, a constant-valued $\tau$ provides at least
some approximation of waiting costs.

It is difficult to infer from \cite{Anderson1993} the exact process by which the above rational decision is made. Two alternative interpretations are possible:
\begin{enumerate}
\item At time $t$ (when production $s_i$ fired), the agent implicitly computes a fixed amount of time $T$ during which it will wait for higher-valued
productions. If no higher-valued production matches during this interval, then $s_i$ is fired; if production $s_j$ matches such that $V_j>V_i$, then the
time $t$ of the last-matched production is updated (to the time $s_j$ matched), and the decision process starts over.
\item At each moment in time $t'\geq t$ (until infinity), the agent will decide, using some binary decision rule $D(t)$,
whether to accept $s_i$ as the best choice, or to wait for a possibly
higher-valued production to match. If at time $t'$ another production $s_j$ matches such that $V_j>V_i$, then the decision process starts over,
with $V_j$ the new value-to-beat.
\end{enumerate}
In fact, these two varying formulations may be reconciliable, if $D(t)$ is chosen somehow to be consistent with $T$.
In ACT-R, the decision rule $D(t)$ is:
\begin{eqnarray}
\begin{array}{ll}
\textrm{\emph{Fire} if} & \int_{V_i}^G (x-V_i) Z_t(x;V_i)dx \leq \tau\\
\textrm{\emph{Wait} otherwise}
\end{array}
\end{eqnarray}
$Z_t(x;V_i)$ is the probability distribution that a production with value $x$ will ever match, some time between $t$ and infinity, and
$\tau$ is the cost of waiting (retrieving a future production), described above. $Z_t$ is defined as (see \cite{Anderson1993}, p. 62, and
\cite{Anderson1990}, p. 215):
\[ Z_t(x;V_i) = \frac{1}{t(G-V_i)} e^{-\frac{1}{t(G-V_i)} (G-x)} \]
This distribution integrates to 1 iff $x \in [- \infty, G ]$, i.e., if future productions are bounded above by the value of
$G$.\footnote{Note that, under this distribution, \emph{higher}-valued productions are given a higher probability of occurring.
It is unclear whether this was intended.}
Note that the probability of a production with value $x$ occurring decreases with time; this was intended by Anderson to model the idea
that higher-valued productions should match earlier than lower-valued productions \cite{Anderson1993}.
It is important to realize that, according to this definition of $Z_t$, $t$ is \emph{not} the time when the production with value $x$ fires.

Applying this decision rule $D(t)$ at each time $t'\geq t$ implicitly results in a time-window $T$ for which the agent will end up waiting
after production $s_i$ fired. Hence, the two alternative formulations posed above are somehow equivalent. However, how to calculate $T$ from
$D(t)$ is unclear.

%
%

\subsection{Selecting from Multiple Productions}
Each production rule has an associated expected value:
\begin{equation}
V = pG - C
\label{eqn:prod_value}
\end{equation}
where $G$ is the reward earned by reaching the current goal, $p$ is the probability that the goal will be
reached if the production is selected, and $C$ is the expected immediate and future costs of selecting
the production.

Suppose a production $i$ with value $V_i$ is matched at time $t$, and suppose $V^*$ is the highest value production that
can ever match the current set of declarative knowledge chunks. (Recall that productions have a latency before they match.)
Then production $i$ will fire (be selected) iff:
\[ E[V^* - V_i] \leq \tau \]
for a fixed threshold $\tau$. This expectation is computed over the distribution $Z_t(V^*)$ which is
the probability of a production firing with value $V^*$. Under ACT-R, $Z$ is dependent on $t$ in order to
``allow for the possibility that later instantiations are less likely to be as good as earlier instantiations''
(\cite{Anderson1993}, p. 62). Hence, production $i$ will fire iff:
\[ \int_V^G (V^*-V) Z_t(V^*) dV^* \leq \tau \]
(Note that $G$ is the maximum value a production could have since $p\in [0,1]$ and $C>0$.)

\section{Learning in ACT-R}
\label{sec:learning}
In ACT-R, learning consists of \emph{creating} new knowledge chunks and production rules in memory,
and in \emph{strengthening} these memories through use. We discuss each category of learning below.

\subsection{Creating New Knowledge Chunks}
Very little is said about how ACT-R models the initial creation of declarative memory: Either a new chunk is ``created as the
encoding of external events'' (\cite{Anderson1993}, p. 70), or a chunk is written to memory as the result of an
executed production rule. The letter case is exemplified in Figure \ref{fig:example_productions}, where the
knowledge chunk (variable $k$) that contains the column currently being processed is updated ($k$ is incremented by 1
in P2).

\subsection{Creating New Production Rules}
The initial creation of productions is somewhat more developed. New production rules are formed through several
processes, including: proceduralization, composition, generalization, and analogy.

\subsubsection{Proceduralization} Production rules often contain variables to refer to
particular values on which the production operates. An example was shown in Figure \ref{fig:example_productions}
for the addition of multi-digit numbers. After performing the same exact task many times, new productions 
can arise in procedural memory which contain the parameter values ``hard-coded'' into them. For instance,
if the addition $36+23$ is performed repeatedly, then the laborious application of the six production rules
could be skipped by creating a new production: ``If the goal is to add 36 and 23, then write 59.'' This is known
as proceduralization and is a form of \emph{compilation} under ACT-R \cite{Anderson1982}. However, ACT-R
proposes no model of when or under what conditions such a proceduralization will occur.

\subsubsection{Composition} If two productions perform problem-solving steps in sequence, then the productions can be
combined. Since applying each production requires some cost, performing both productions in one step can potentially
save computational resources such as time and/or effort. For an example of composition, if P1 (not the one
in Table \ref{tbl:example_productions}) has the effect of multiplying both sides of the equation $\frac{4}{5}x=1$
by 5, and P2 divides both sides by 4 (to solve for $x$), then a new production -- call it P3 -- could combine
these steps and multiply both sides by $\frac{5}{4}$.

\subsubsection{Generalization} Generalization occurs when many similar production rules have already been stored, from which a more general
rule can be learned inductively. For example, a child may already know that the plural of ``cat'' is ``cats,'' and
that the plural of ``dog'' is ''dogs.'' From these rules, he/she may infer that a plural can be formed
by appending an ''s'' to the singular. The astute reader will note that there are many exceptions to this rule -- for instance,
``ox'' and ``oxen'', or ``octopus'' and ``octopodes.''\footnote{The proper English plural form of ``octopus'' is either ``octopodes'' or ``octopuses.''}
This necessitates what is called \emph{discrimination}, in which additional production rules are learned to handle the exceptions.

\subsubsection{Analogy} When presented with a new problem for which no solution (in the form of production rules) is known,
a human can reason by analogy from previously seen example solutions to similar problems. For example, having encountered
the programming problem in which a variable is to be increased by 4, whose solution (in C) is \begin{tt}x += 4\end{tt},
he/she may reason that the solution to the similar task of multiplying a variable by 4 might be \begin{tt}x *= 4\end{tt},
assuming the syntax for multiplication (\begin{tt}*\end{tt}) was already known. Analogy-based learning requires 
that the learner bew able to create a mapping from the known example to the present task. In the case above, the mapping
was from addition to multiplication. The new production rule is created by applying the same mapping to the solution
production of the example problem, i.e., ``...Then write x += 4'' is mapped to ``...Then write x *= 4.''

Analogy may also occur even if (perhaps inefficient) production rules already exist which can solve the current problem.
According to Anderson, this could potentially be modeled under ACT-R by assuming the existence of a meta-production -- analogy --
which models the value of learning new productions for future use.

\subsection{Strengthening Existing Chunks}
In ACT-R, the strength in memory of a knowledge chunk is called \emph{activation}. A chunk can gain activation through \emph{use}: a
chunk is \emph{used} when it matches against some production rule that fires. This definition is also consistent with the idea
of memory strength increasing through practice by introducing a trivial ``recall'' production:
by defining the current goal to be ``Practice $i$,'' and defining a production as
``If the goal is to practice $i$, then recall $i$.''

Anderson posits that the \emph{activation} $A(t)$ of a knowledge chunk is the log-odds that, at time $t$, it will match to some production rule that ends up firing.
\footnote{In fact, he defines activation in two conflicting ways: in \cite{Anderson1993}, p. 64, chunk activation is the log-odds that the 
chunk matches a production that \emph{fires}; in \cite{Anderson1993}, p. 50, however, activation is merely the log-odds of matching \emph{any} production,
not necessarily the one that fires. This distinction is crucial.} The activation $A(t)$ of a particular chunk is defined as:
\begin{equation}
A(t) = B(t) + \sum_j w_j s_{ij}
\label{eqn:activation}
\end{equation}
where $B(t)$ is the \emph{base activation} of the knowledge chunk at time $t$, and the summation is the associative strength of the
chunk dependent on related chunks, defined as ``the elements in the current goal chunk and the elements currently being
processed in the perceptual field'' (\cite{Anderson1993}, p. 51). Each $w_j\in[0,1]$ ``reflects the salience or validity of [related] element $j$.''
The $s_{ij}\in[-\infty,+\infty]$ are ``the strengths of association to $i$ from elements $j$ in the current context.''

The base activation $B(t)$ is increased whenever it is ``used,'' either through practice (``learning events'') or when matched to a production rule.
\begin{equation}
B(t) = \ln \sum_k t_k^{-d} + B
\label{eqn:base_activation}
\end{equation}
The decay rate $d$ is defined in the ACT-R literature in various ways:
\label{sec:decay_rate}
\begin{itemize}
\item In \cite{Anderson1993}, the decay rate $d$ is a constant for all $k$.
\item In \cite{AndersonSchooler1991}, the decay rate of activation for the particular learning event $k$ is defined as
$d_k = \max \{ d_1, b(t_{k-1} - t_k)^{-d_1} \}$\footnote{In \cite{AndersonSchooler1991}, the base was actually $t_k - t_{k-1}$, but since older events
have longer decay times $t$, this would result in a negative base, and hence a complex number for the decay, which was presumably unintended.}
where $d_1$ (a constant) is the decay rate of the first learning event for the chunk, and $b$ is a constant. The difference $t_{k-1} - t_k$ expresses
the \emph{spacing effect} --
the notion that tight spacing of learning events results in lesser activation gains than longer spacing.
\item In \cite{PavlikAnderson2008}, $d_k = ce^{m_{k-1}} + \alpha$, where $c$ and $\alpha$ are constants, and $m_{k-1}$ is the activation of the chunk at the
time of the previous learning event ($k-1$). This definition also expresses the spacing effect since, if the activation was already high at the previous
learning event $k-1$, then the decay rate of the $k$th learning event will also be high.
\end{itemize}

\subsection{Learning Associative Weights}
The activation $A_i(t)$ of chunk $i$ is also afffected by the presence of other chunks in the current \emph{context}.
The notion of context in ACT-R is not clearly defined, but it is described roughly as 
``the elements in the current goal chunk and the elements currently being processed in the perceptual field'' (\cite{Anderson1993}, p. 51).
Presumably, the elements of the goal chunk refer to those chunks to which the goal chunk is \emph{connected} (in the sense of Section \ref{sec:decl_proc}).

Suppose chunk $j$ is in the current context. Then its effect on $A_i(t)$ is as follows: if $i$ is connected to $j$, then the log-odds of
chunk $i$ matching to a production that fires (i.e., $i$'s activation) is increased by $w_j s_{ij}$. The $w_j$ are barely defined at all,
except that each $w_j\in[0,1]$, and that it ``reflects the salience or validity of [related] element $j$.'' \cite{FumStocco2004} suggests
that the $w_j$ express ``attentional weighting, i.e., the variable degree of attention different individuals are able to dedicate to the elements
of the current context.''

The $s_{ij}\in[-\infty,+\infty]$ are ``the strengths of association to $i$ from elements $j$ in the current context'' (\cite{Anderson1993}, p. 51).
They are weights that are learned based on the history of chunk $i$ matching a production that fires with or without the accompanying presence
of chunk $j$ in the current context. More precisely, $s_{ij}$ represents the log-likelihood ratio term
\[ \log \frac{p(C_j\ |\ N_i)}{p(C_j\ |\ \overline{N_i})} \]
where $C_j$ is the event that chunk $j$ is in the current context and $N_i$ is the event that $i$ matches a production that fires. $s_{ij}$
can be approximated as
\[ s_{ij} = \log \frac{p(C_j\ |\ N_i)}{p(C_j\ |\ \overline{N_i})} \approx \log \frac{p(C_j\ |\ N_i)}{p(C_j)} = \log \frac{p(N_i\ |\ C_j)}{p(N_i)} \]
The justification for this approximation given in \cite{Anderson1993} is that knowing the outcome of $\overline{N_i}$ cannot substantially affect
the probability of event $N_i$ since there are presumably very many chunks in all of declarative memory.
The two terms in the fraction, $p(N_i\ |\ C_j)$ and $p(N_i)$, can be estimated empirically based on the history of productions that fired, and
the chunks that matched them, and the context at each moment a production fired. In \cite{Anderson1993}, a prior probability for both of these terms
was also employed.

\subsection{Learning Production Success Probability and Production Costs}
In order to estimate the value of a production $V=pG-C$, the terms $p$ and $C$ must both be computed, where $p$ is the probability
that the production will lead to the goal being achieved, and $C$ is the expected immediate and future cost of firing this production.
Probability $p$ can be decomposed into the probability that the production $s$ itself will succeed, i.e., have its intended affect, multiplied
by the probability that the goal will be reached conditional on production $s$ succeeding. The first probability, which is called $q$
in \cite{Anderson1993}, can be estimated by updating a Beta distribution every time that a production either succeeds or fails. The second quantity,
called $r$, is estimated by ACT-R heuristically in one of two different ways: from the cost already incurred in achieving the goal (the assumption
is that, the higher the total cost already spent, the lower the probability should be), or from the similarity between the current state and
the goal state.

The notion of a production ``succeeding'' and the real-world ``costs'' involved in a production are not thoroughly developed in ACT-R. As stated
by the author himself in \cite{Anderson1990}, the proposed methods of estimating $p$ and $C$ are not optimal, but rather only plausible. We do
not review them in detail in this tutorial. More conventional methods, from the perspective of the machine learning community, for estimating
these quantities are available in the reinforcement learning literature for Markov processes.

\subsection{Strengthening Existing Productions}
Analagous to \emph{activation} for knowledge chunks, the measure of memory strength of a production rule is the \emph{production strength}, $S$.
In \cite{Anderson1993}, $S$ is supposed to be the log-odds that the production rule will fire,  and it is defined as:
\begin{equation}
S(t) = \ln \sum_k t_k^{-d} + B
\label{eqn:prod_strength}
\end{equation}
where the $k$ indexes events in which the production fires (\cite{Anderson1993}, p. 293), and $B$ is a constant.
(It is unclear from \cite{Anderson1993} if the $B$ in \ref{eqn:prod_strength} is the same $B$ as in \ref{eqn:base_activation}.)

\subsubsection{Cautionary Note} Equation \ref{eqn:activation} supposedly models the log-odds that a particular chunk will match the production
that fires, and \ref{eqn:prod_strength} models the log-odds of a particular production rule firing. However, these probabilities are not independent:
for instance, if chunk $i$ is the sole chunk referenced by production rule $s$, and if $A_i(t)=-\infty$ (i.e., the chunk has no chance of being matched to the rule
that fires), then $S(t)$ must also equal $-\infty$. Moreover, since production firing is modelled as a competition between multiple production
rules that fired, the log-odds of a particular production firing cannot be calculated independently of other productions.
For both the above reasons, the associated equations cannot truly equal the log-odds that they are claimed to represent.

\subsection{Latency of Production Rule Matching}
Learning can also decrease the latency involved in a production rule matching. The \emph{latency} of a production rule $s$ is
defined as the period between the first moment that all of the knowledge chunks in the pre-condition of $s$ were in memory,
to the time when $s$ actually matches. This latency is modeled as:
\begin{equation}
L(t) = \sum_i Be^{-b(A_i(t)+S(t))}
\label{eqn:latency}
\end{equation}
where the summation is over the required knowledge chunks for the production, $A_i(t)$ is the ``activation'' (a measure of strength
in memory, described below) of the knowledge chunk at time $t$, and $S(t)$ is the strength of the production itself at time $t$. $B$ and
$b$ are constants.

\section{Memory}
\subsection{Relationship of Latency to Probability of Recall}

The probability of recalling a knowledge chunk is 
\[ P(t) = \frac{1}{1+ e^{-(A(t)-\tau)/s}} \]
which means that a knowledge chunk will tend not to be remembered if its activation $A(t)$ is below $\tau$. This is consistent
with ACT-R because certain productions will never fire if their latency is too long.

\section{The Spacing Effect}
\label{sec:spacing}
The spacing effect is a psychology phenomenon whereby probability of remembering an item at test-time can be increased by scheduling practice
sessions such that a substantial time difference exists between successive sessions. Intuitively, if one had just drilled a particular item a few
seconds ago, then it will make little difference to the person's memory if he/she immediately drills it again. The spacing effect can be modeled
in ACT-R in at least two different ways, as explained in Section \ref{sec:decay_rate}.

\section{The Power Laws of Learning}
\label{sec:power_laws}
\subsection{Latency}
Though much touted for supposedly modeling the power law of learning, ACT-R in fact exhibits only a tenuous relationship with this phenomenon.
One form of the power law of learning is that the latency of a correctly retrieved declarative memory chunk $L$ decays as a power function,
which has the general form $L(t) = At^{-k}$ for constants $A$ and $k$
assuming that the number of practice sessions grows with $t$. However,
in ACT-R, in order for this relationship to hold, one must assume that the decay rate $d$ of multiple practice sessions is constant
(which then destroys the modeling of the spacing effect), and also that
the time between practice sessions $t_{k-1} - t_k$ is likewise some constant $\Delta t$. One must also consider the latency of
a knowledge chunk in isolation from any particular production, as in \cite{PavlikPressonAnderson2007} --
in the ACT-R definition in \cite{Anderson1993}, however, latency is
only explicitly defined for the instantiation of a production rule (matching of its pre-condition to chunks); hence, strictly speaking,
this too is an approximation.

Under these assumptions, the expected latency 
of recall of a knowledge chunk is \emph{approximately} modeled by a power function. (The proof below is
adapted from \cite{AndersonFinchamDouglass1999}).
The latency of recall of a particular chunk as defined in \cite{PavlikPressonAnderson2007} is:
\begin{eqnarray*}
L(t) = Fe^{-A(t)} + C
\end{eqnarray*}
where $A(t)$ is the activation of the chunk at time $t$, and $F$ and $C$ are constants, such that $C$ represents
some fixed time cost of retrieval of the chunk.
Since the only component of $A(t)$ that depends on $t$ is $B(t)$, let us ignore these terms as well (since they can
be folded into $F$) -- these terms are the associative strength term and the constant $B$ term in Equation \ref{eqn:base_activation}.
Then,
\begin{eqnarray*}
L(t) &=& F e^{-B(t)}\\
     &=& F e^{- \ln \sum_{k=0}^K t_k^{-d}}\\
     &=& F \frac{1}{\sum_{k=0}^K t_k^{-d}}
\end{eqnarray*}
Now, since the learning events indexed by $k$ were assumed to be equally spaced in time, we can rewrite this as:
\begin{eqnarray*}
L(t) &=& F \frac{1}{\sum_{k=0}^K ((K-k) \Delta t)^{-d}}\\
     &=& F \frac{(\Delta t)^d }{\sum_{k=0}^K (K-k)^{-d}}\\
     &=& F \frac{(\Delta t)^d }{\sum_{k=0}^K k^{-d}}\\
     &\approx& F \frac{(\Delta t)^d }{\int_{k=0}^K k^{-d}dk}\\
     &=& F \frac{(\Delta t)^d }{K^{1-d}/(1-d)} \\
     &=& F (\Delta t)^d (1-d) K^{d-1}
\end{eqnarray*}
The number of learning events $K$ is a function of $t$: $K(t)=\frac{t}{\Delta t}$. Hence:
\begin{eqnarray*}
L(t) &=& F (\Delta t)^d (1-d) \left( \frac{t}{\Delta t}\right)^{d-1}\\
     &=& G t^{d-1}
\end{eqnarray*}
for some constant $G$.


{\small
\bibliographystyle{ieee}
\bibliography{./paper}
}

\end{document}